\pdfoutput=1

\documentclass[a4paper, 11pt]{article}
\usepackage[margin=2cm]{geometry}
\usepackage[english]{babel}
\usepackage[utf8x]{inputenc}
\usepackage{graphicx}
\usepackage{hyperref}
\usepackage{placeins}
\usepackage{float}
\usepackage{subfigure}
\hypersetup{colorlinks=true,citecolor=black,filecolor=black,linkcolor=black,urlcolor=black}

\usepackage[numbers,square]{natbib}

\begin{document}

	\title{
	\vspace{4.0cm}
	\Huge Deep Learning for Medical Image Segmentation \\
    \vspace{1.5cm}
	}
	
	\author{\Large \href{mailto:matthew.lai14@imperial.ac.uk}{Matthew Lai} 
	\vspace{1.5cm}}
	
	\date{
	\large Supervisor: Prof. Daniel Rueckert \\
	\vspace{1.5cm}
	Apr 29, 2015
	}

	\maketitle
	\setlength{\parindent}{0pt}

\vspace{1cm}
\begin{abstract}
\thispagestyle{empty}
This report provides an overview of the current state of the art deep learning architectures and optimisation techniques, and uses the ADNI hippocampus MRI dataset as an example to compare the effectiveness and efficiency of different convolutional architectures on the task of patch-based 3-dimensional hippocampal segmentation, which is important in the diagnosis of Alzheimer's Disease. We found that a slightly unconventional "stacked 2D" approach provides much better classification performance than simple 2D patches without requiring significantly more computational power. We also examined the popular "tri-planar" approach used in some recently published studies, and found that it provides much better results than the 2D approaches, but also with a moderate increase in computational power requirement. Finally, we evaluated a full 3D convolutional architecture, and found that it provides marginally better results than the tri-planar approach, but at the cost of a very significant increase in computational power requirement.

\end{abstract}
	
	\newpage
	\tableofcontents
	\newpage
	\listoftables
	\newpage
	\listoffigures
	\newpage
	
\section{Introduction} 

Deep learning techniques have been applied to a wide variety of problems in recent years \cite{arel2010deep} - most prominently in computer vision \cite{krizhevsky2012imagenet}, natural language processing \cite{collobert2008unified}, and computational audio analysis \cite{lee2009unsupervised}. In many of these applications, algorithms based on deep learning has surpassed the previous state-of-art performance. At the heart of all deep learning algorithms is the domain-independent idea of using hierarchical layers of learned abstraction to efficiently accomplish high level tasks \cite{arel2010deep}.

This report first provide an overview of traditional artificial neural network concepts introduced in the 1980s, before introducing more recent discoveries that made training deep networks practical and effective. Finally, we present the results of applying multiple deep architectures to the ADNI hippocampus segmentation problem, and comparing their classification performances and computational power requirements.

	
\section{Traditional Neural Networks} 
\label{sec:traditional_neural_networks}

Artificial neural networks (ANN) are a machine learning technique inspired by and loosely based on biological neural networks (BNN). While they are similar in the sense that they both use a large number of identical and linked simple computational units to achieve high performance on complex tasks, modern ANNs have been so heavily optimized for efficient implementation on electronic computers that they bear little resemblance to their biological counterpart. In particular, time-dependent integrate-and-fire mechanism in BNNs have been replaced by steady state values representing frequency of firing, and most ANNs also have vastly simplified connection architectures that allow for efficient propagation. Most current ANN architectures don't allow loops in connections (with the notable exception of recurrent neural networks which uses loops to model temporal correlations \cite{williams1989learning}), and also don't allow connections to be made and broken during training (with the notable exception of evolving architectures based on genetic algorithms \cite{leung2003tuning,belew1990evolving,schaffer1992combinations}). Unless otherwise specified, all further references to neural networks (NN) in this report refer to artificial neural networks.

\subsection{Network Architectures}
\label{sec:network_architectures}
In a typical neural network, nodes are placed in layers, with the first layer being the input layer, and the last layer being the output layer. The input nodes are special in that their outputs are simply the value of the corresponding features in the input vector.

For example, in a classification task that has a 3-dimensional input (x, y, z) and a binary output, one possible network design is to have 3 input nodes, and 1 output node. The input and output layers are usually considered fixed in network design.

With only an input layer and an output layer, with all input nodes connected to all output nodes, the network essentially implements a matrix multiply, or a linear transformation. This type of networks can solve simple problems where the feature space is linearly-separable. However, for linearly-separable problems, simpler techniques such as linear regression or logistic regression can usually achieve similar performance, with the only difference being training methods. 

Most modern applications of neural networks use one or more hidden layers – layers that sit between the input layer and the output layers, to allow the network to model non-linearity in the feature space. The number of hidden layers and the number of hidden nodes in each layer are hyper-parameters that are not always easy to determine. While some rules-of-thumb have been proposed, they are still, for the most part, determined by trial-and-error. The risk of using too-small a network is that it may not have enough representative power to model all useful patterns in the input (high bias), while the risk of using too-large a network is that it may overfit the data, and start modeling noise in the training set (high variance). It is usually better to err on the side of larger networks, because many effective techniques exist to combat overfitting, as will be detailed in later sections of the report. Using network size to limit overfitting is error-prone, time-consuming, and not very effective.

Methods have been proposed for automatic hyperparameter tuning, such as evolving cascade networks \cite{schetinin2003learning}, which trains a large network through an iterative process, by first starting with a minimal network, and in each iteration, train a few "candidate" networks that have more nodes in different layers, and keeping the best. There are also tuning methods based on genetic algorithms with links turned on and off by each bit in the genes \cite{leung2003tuning,belew1990evolving,schaffer1992combinations}. However, these methods have not seen widespread adoption, due to the large increase in training time, and marginal benefits when overfitting is avoided using other methods than limiting network size.

It has been proven that a network with 1 hidden layer can approximate any continuous (in feature space) function to any accuracy, and a network with 2 hidden layers can approximate any function to any accuracy \cite{hornik1989multilayer,leshno1993multilayer}. Given infinite computational power, memory, and training set, there is theoretically no reason to go above 2 hidden layers. However, as will be explained later in the report, it is much more efficient to solve complex problems using a deeper network than one with only 2 hidden layers.


\subsection{Network Nodes and Activation Functions}
\label{sec:network_nodes_and_activation_functions}

In a neural network, each node (besides input nodes) has one or more scalar inputs, and one output. Each link between nodes have a scalar weight, and each node has a bias to shift the point of activation.

\begin{equation}
\label{eq:node_output}
f(\sum{w_i * x_i} + b)
\end{equation}

The output of each node is computed as shown in Equation~\ref{eq:node_output}, where $x_i$'s are inputs to the node, $w_i$'s are the weights of the associated link, $b$ is a bias associated with the node, and $f(x)$ is a function associated with the node, known as the activation function.

There are a few activation functions in widespread use. For output nodes in regression networks, a linear activation function (eg. $y = x$) is most commonly used to give these networks a range of all real numbers. For output nodes in classification networks, the softmax function (exponential normalization) is often used to transform the outputs into something that can be interpreted as a probability distribution.

For hidden nodes, the traditional choices are hyperbolic tangent ($y = tanh(x)$) and the logistic function ($y = \frac{1}{1+e^{-k(x - x_0)}}$). Both these functions are designed to satisfy 3 conditions -
\begin{itemize}
\item Differentiable everywhere
\item Monotonic
\item Non-linear
\end{itemize}

It was believed that these properties are essential for an activation function. 

The differentiability property is important because we must be able to take the derivative of the function at any point during training using gradient-based methods. It is not necessary for networks trained using non-gradient-based methods such as genetic algorithm.

The monotonicity property is important because if the activation function is not monotonic, it will introduce additional local minimums in the parameter space, and impede training efforts.

Non-linearity is important because otherwise the network will lose the ability to model non-linear patterns in the training set. Non-linearity is achieved using saturation in this case, with the hyperbolic tangent function saturating at $y = -1$ and $y = 1$, and the logistic function saturating at $y = 0$ and $y = 1$.

In practice, although the logistic function is more biologically plausible, hyperbolic tangent usually allows faster training since being linear around $0$ means nodes will not start training in saturation (which would make training much slower) even if inputs are zero or negative \cite{glorot2011deep}.

\subsection{Training Neural Networks}
\label{sec:training_neural_networks}

Training algorithms for neural networks fall into two major categories - gradient-based and non-gradient-based. This report focuses on gradient-based methods as it is much more commonly used in recent times, and usually converges much faster as well.

As mentioned in Section~\ref{sec:network_nodes_and_activation_functions}, each node in the network has a weight associated with each incoming link, and a scalar bias. The weight and bias of a node are the parameters of the node. If we concatenate the weights and biases of all nodes in a network into one vector $\theta$, it completely defines the behaviour of a network (for a given set of hyper-parameters, ie. network architecture).

\begin{equation}
\label{eq:network_output}
y = f(\theta, x)
\end{equation}

If the set of hyper-parameters (network architecture) is encoded into a function $f()$, we can define the output of the network as shown in Equation~\ref{eq:network_output}, where y and x are the output and input vectors respectively.

The goal of the training process, therefore, is to find a $\theta$ such that $f(\theta, x)$ approximates the function we are trying to model. In other words, given a set of inputs and their desired outputs, we are trying to find a $\theta$ that minimizes the difference between the desired outputs and network outputs, for all entries in the training set. For that, a measurement of error is needed.

\begin{equation}
\label{eq:mse_error}
E(\theta, Ts) = \frac{1}{2} \sum_{(x_i, y_i) \in Ts} (y_i - f(\theta, x_i))^2
\end{equation}

One such error measure is mean-squared-error (MSE), and it is the most commonly used error measure. This is given in Equation~\ref{eq:mse_error}, where Ts is the training set, $x_i$ and $y_i$ are the input and desired output of a training pair, N is the number of entries in the training set, and $g(x_i)$ is the network output.

Our goal is to minimize $E(\theta, Ts)$ given $Ts$. For very small networks, it may be feasible to do an exhaustive search to find the point in parameter space where the mean-squared-error is minimized, but for networks of reasonable sizes, an exhaustive search is not practical. Gradient descent is the most commonly used optimisation algorithm for neural networks.

In gradient descent, $\theta$ is first initialized to a random point in the parameter space. Weights and biases are typically drawn in a way that keeps most nodes in the linear region at the beginning of training. One popular method is to draw from the uniform distribution $(-\frac{a}{\sqrt{d_{in}}}, \frac{a}{\sqrt{d_{in}}})$, where $a$ is chosen based on the shape of the activation function (where it starts to saturate), and $d_{in}$ is the number of inputs to the node \cite{yam2000weight}.

\begin{equation}
\label{eq:gradient_descent}
\Delta \theta = -L \frac{\partial E(\theta, Ts)}{\partial \theta}
\end{equation}

After initialization, the gradient descent algorithm performs a walk in the parameter space, guided by the gradient of the error surface. In its simplest form, in each iteration, the algorithm takes a step in the opposite direction of the gradient, with the step size proportional to the magnitude of the gradient, and a fixed learning rate. This is shown in Equation~\ref{eq:gradient_descent}, where $L$ is the learning rate, and all other symbols are as previously defined.

\begin{equation}
\label{eq:error_function_d}
\frac{\partial E(\theta, Ts)}{\partial w_{kj}} = \frac{\partial E}{\partial y_j} \frac{\partial y_j}{\partial x_j} \frac{\partial x_j}{\partial w_{kj}}  
\end{equation}

The partial derivative of the error function for each parameter is as shown in Equation~\ref{eq:error_function_d} after applying chain rule, where $x_j$ is the result of the summation (input to activation function), $y_j$ is the output of the activation function, and $w_{kj}$ is the weight we are examining.

The only difference between the output layer and hidden layers is that hidden layers do not really have an error. However, we can still derive error terms for them by calculating its contribution to the input to the activation function of the next node. This is equivalent to another application of chain rule, since the input to the activation function is simply the sum of the contributions from each node in the previous layer. It is convenient to do this propagation backwards, from the final error, multiplying by the derivative of the activation function each time, before "assigning blame" of the error proportionally to the weights connecting previous layer nodes to the node we are examining. This is the basis of back-propagation, and this is why we require the activation function to be differentiable. A detailed derivation is omitted here for brevity.

Many variants of gradient descent have been proposed, each with different performance characteristics. The one in most popular use is gradient descent with momentum, where instead of calculating the gradient each time and use that as the step, we combine it with a fraction of the weight update of the previous iteration. This allows faster convergence in situations where the gradient is much larger in some dimensions than others (eg. down the bottom of a valley) \cite{pearlmutter1991gradient}.

Another common variation is learning rate scheduling - changing the learning rate as training progresses. The goal is to get to somewhere close to a local minimum quickly, then slow down to avoid overshooting. This idea is taken further in resilient back-propagation (RPROP), where only the sign of the gradient is used. In RPROP, each weight has an independent learning rate, that is increased (usually multiplied by 1.2) if the sign of the gradient has not changed from the previous iteration, and reduced (usually by a factor of 0.5) if the gradient has changed signs \cite{riedmiller1992rprop}. This allows all weights to train at close to their optimal learning rate, and eliminates the learning rate parameter that must be manually tuned in other gradient descent variants. An initial learning rate still needs to be chosen, but it doesn't significantly affect training time or result \cite{riedmiller1992rprop}.


\subsection{Regularization}
\label{sec:regularization}

When the training set is limited in size as it is usually, it is dangerous to train without any constraint since the network will eventually start to model noise in the training set, and be too specialized to generalize beyond the training set.

\begin{equation}
\label{eq:error_with_l2_reg}
E(\theta, Ts) = \frac{1}{2} \sum_{(x_i, y_i) \in Ts} (y_i - f(\theta, x_i))^2 + \lambda |\theta|^2
\end{equation}

One popular way to combat overfitting is regularization - the idea of encouraging weights to have smaller values. The most common form of regularization is L2 regularization, where the L2 norm of $\theta$ (the parameter vector) is added to the error function, as shown in Equation~\ref{eq:error_with_l2_reg}, where $\lambda$ is a parameter that controls the strength of regularization, and it needs to be tuned. If $\lambda$ is too low, the network would overfit. If $\lambda$ is too high, the network would underfit.

Other norms such as $L1$ and $L\frac{1}{2}$ are also used, with different effects. They will be explored in the discussion of deep networks later in the report.



\section{Deep Learning} 
\label{sec:deep_learning}

\subsection{Why Build Deep Networks?}
\label{sec:why_deep}

As mentioned earlier in Section~\ref{sec:network_architectures}, a neural network with 2 hidden layers is already theoretically a universal function approximator capable of approximating any function, continuous or not, to any arbitrary accuracy. In light of that, it may seem pointless to pursue networks with more hidden layers.

The main benefit of using deep networks is node-efficiency - it is often possible to approximate complex functions to the same accuracy using a deeper network with much fewer total nodes compared to a 2-hidden-layer network with very large hidden layers. Besides computational benefits, a model with a smaller degree of freedom (number of parameters) requires a smaller dataset to train \cite{schwarz1978estimating}, and size of the training set is often a limiting factor in neural network training.

Intuitively, the reason for a smaller and deeper network to be more effective than an equally sized (in total nodes) shallower network is that a deep network reduces the amount of redundant work done.

As an example, suppose we are given bitmap images containing triangles, rectangles, pentagons, and hexagons, and our task is to count the number of each shape in each image. In the case of a deep network, the first layers can perform the low level task of calculating image gradients and identifying lines in the image, and transform it to a more convenient form. Higher layers can then perform classification and counting from the simpler representation.

On the other hand, if we have a shallow network, the low level tasks would have to be performed multiple times, since there is little cross-feeding of intermediate results. This would result in a lot of redundant work done.

With very deep networks, it is possible to model functions that work with many layers of abstraction - for example, classifying the gender of images of faces, or the breed of dogs. It is not practical to perform these tasks using shallow networks, because the redundant work done is exponential to the number of layers, and an equivalent shallow network would require exponentially more computational power, and exponentially larger training sets, neither of which are usually available.


\subsection{Vanishing Gradients}
\label{sec:vanishing_gradients}

The fact that deeper networks are more computationally efficient has been known for a very long time, and deep networks have been attempted as early as 1980 \cite{fukushima1980neocognitron}. In 1989, LeCun et al. successfully applied a 3-hidden-layer network to ZIP code recognition. However, they were not able to scale to higher number of layers or more complex problems due to very slow training, the reason of which was not understood.

In 1991, Hochreiter identified the problem and called it "vanishing gradients" \cite{hochreiter1991untersuchungen,bengio1994learning}. Essentially, what's happening is as errors are propagated back from the output layer, it's multiplied by derivatives of the activation function at the point of activation. As soon as the propagation gets to a node in saturation (where derivative is close to $0$), the error is reduced to the level of noise, and nodes behind the saturated node train extremely slowly. No effective solution to the problem was found for many years.

Researchers continued to try to build deep networks, but with often disappointing performance.

\subsubsection{Solution 1: Layer-wise Pre-training}
\label{layer_wise}

In 2007, Hinton proposed a solution to the problem, and started the current wave of renewed interest in deep learning. The idea is to first train each layer in an unsupervised and greedy fashion, to try to identify application-independent features from each layer, before finally training the entire network on labeled data \cite{hinton2007learning}.

The way he implemented the idea is with a generative model, in the form of a restricted Boltzmann machine (RBM), where each layer contains a set of binary variables with associated probability distributions, and each layer is trained to predict the previous layer using an algorithm known as contrastive divergence \cite{hinton2007learning}.

Another idea in the same vein is to use a neural network to reproduce the original input \cite{vincent2008extracting}, which allows the reuse of all the neural network techniques that have already been developed (unlike for RBM). 

The idea is to first start with just the input layer and one hidden layer, and train the network (using standard back-propagation gradient descent) to produce the original input, essentially training the network to model the identity function \cite{vincent2008extracting}. If the hidden layer has fewer nodes than the input layer, the output of the hidden layer is then a compressed and more abstract representation of the input. This process is repeated to train each hidden layer, with the output of the previous layer as the input \cite{vincent2008extracting}. The final network can then be trained with the actual output layer using standard back-propagation from labeled data. This works around the vanishing gradient problem because when the final back-propagation is performed, most of the earlier layers are already trained to provide application-independent abstractions.

In modern implementations, some artificial noise is often injected into the input of autoencoders to encourage robustness in the autoencoders, by testing their ability to reconstruct clean input from partially-corrupted input \cite{vincent2008extracting}. These autoencoders are known as denoising autoencoders. Denoising autoencoders perform similarly to systems based on RBMs and their stacked variant - Deep Belief Networks (DBN) \cite{vincent2008extracting}.


\subsubsection{Solution 2: Rectified Linear Activation Units}
\label{relu}

\FloatBarrier
\begin{figure}[ht!]
    \centering
    \includegraphics[scale=0.5]{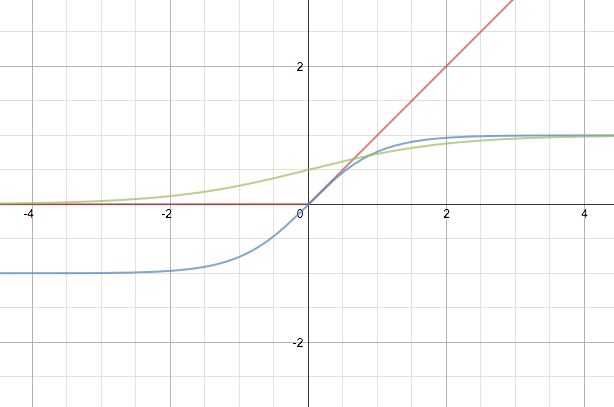}
    \caption{Activation functions - red: ReLU, blue: tanh, green: logistic}
    \label{fig:activation_functions}
\end{figure}

In 2011, a simpler solution to the vanishing gradients problem was proposed by Glorot et al. Their solution to the vanishing gradients problem is to simply use an activation function that does not reduce the error as it's propagated back \cite{glorot2011deep}.

The proposed function is the rectified linear activation function (ReLU), $y = max(0, x)$ \cite{glorot2011deep}. See Figure~\ref{fig:activation_functions} for a comparison of the three activation functions. 

In the activated region ($x > 0$), the derivative is $1$, and in the deactivated region ($x <= 0$), the derivative is $0$. It doesn't change error signals as they are passed through.

At first glance it seems like a strange choice for an activation function, for 2 reasons -
\begin{itemize}
\item It is not differentiable at zero
\item It is unbounded on the positive side
\end{itemize}

The non-differentiability at zero is inconsequential, since neural networks work in real numbers, it is highly unlikely that it will be at exactly $x = 0$ at any time. In practice, we define the derivative at zero to either be $0$ or $1$, and it doesn't have any real world effect.

The unbounded nature of the function on the positive side is more problematic, because they can result in very large activations in later layers, which may cause numerical problems. This can be solved using $L1$ regularization, which not only limits the magnitude of weights, but also enforces sparsity \cite{glorot2011deep}.

\FloatBarrier
\begin{figure}[hb]
\hfill
\subfigure[Distributed weights]{\includegraphics[scale=0.5]{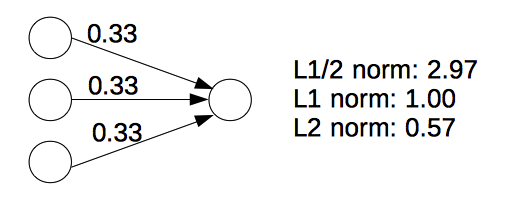}}
\hfill
\subfigure[Sparse weights]{\includegraphics[scale=0.5]{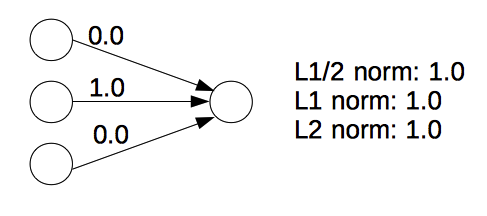}}
\hfill
\caption{Example of different norms}
\label{fig:norms}
\end{figure}
\FloatBarrier

A sparse network is one where for any given input, only a small subset of nodes will be activated. The fact that $L1$ and $L\frac{1}{2}$ regularization encourage sparsity is in contrast to $L2$ regularization which discourages sparsity. $L2$ regularization encourages the desired output to be constructed from many small weights (which would have a lower $L2$ norm) instead of one large weight, which would have a higher $L2$ norm, but equal $L1$ norm, and lower $L\frac{1}{2}$ norm, as shown in Figure~\ref{fig:norms}. In essence, $L1$ and $L\frac{1}{2}$ regularization encourage nodes to be independent from one another, and not co-evolve with others. This can result in more efficient usage of available nodes, and also have computational advantages in networks using activation functions with a hard $0$ saturation (such as the rectified linear activation) \cite{glorot2011deep}, since if a node's output is $0$, it does not need to be broadcasted to nodes in the next layer.

With rectified linear activation, $L1$ regularization, and no pre-training, Glorot et al. reports achieving slightly superior or similar performance to previous results based on pre-training \cite{glorot2011deep}.

One possible reason for the increase in performance is more efficient use of nodes. With a pre-training system, nodes are trained to extract all discriminating patterns from the input data, even if some of those patterns are irrelevant for the task at hand. Without pre-training, those irrelevant patterns would not be encoded, thus freeing up the nodes for more important patterns for the task.

However, he also identified one situation where pre-training is still advantageous - for semi-supervised problems. In a semi-supervised problem, a large training set is available, but only a small subset is labeled. In this case, starting with pre-training using the entire dataset before training the whole network with the labeled subset can result in a network that generalizes better \cite{glorot2011deep}.

For purely-supervised problems, most researchers have abandoned the pre-training idea, and adopted ReLU + $L1$ regularization instead.

There is not sufficient research to decide whether ReLU is also superior to traditional activation functions in shallower networks. Most shallow networks still use hyperbolic tangent or logistic activation, because they are not affected by the vanishing gradient problem.



\subsection{DropOut}
\label{sec:dropout}

In 2012, Hinton et al. proposed a technique that greatly improves network performance in cases where there is limited training data \cite{hinton2012improving}. The idea stems from the observation that when the training set is small, there will be many possible models that will perform well on the training set, but only some will perform well on the testing set.

The traditional solution to this is to train an ensemble of networks from different initialization and/or subsets of the training set, then combine their outputs to produce the final output (often by averaging, or voting) \cite{zhou2002ensembling}. This approach is known to work well in improving model performance, but it is often not computationally feasible in deep learning, where training a single network can take many hours or days even on a fast GPU. Many trained neural networks are also used in real-time applications, and using an ensemble would make the model many times slower to evaluate.

Hinton et al.'s proposal, termed "DropOut", is to randomly de-activate 50\% of the nodes of a network on each training iteration. A disabled node would not participate in forward propagation (where they would output 0), and would block any error signal from propagating through the node during back-propagation. When training is done, all nodes are re-enabled, but all weights are halved to maintain the same output range \cite{hinton2012improving}.

They were able to achieve similar results as ensembles of large numbers of networks, with only about $2 \times$ the computational power requirement of a single network. They hypothesized that this is because disabling 50\% of nodes at random on each iteration forces nodes to independently evolve, instead of co-evolving with others. Co-evolution is not optimal because, for example, some nodes can evolve to correct mistakes made by other nodes, instead of modeling useful patterns. In DropOut, nodes cannot assume other nodes exist, and are forced to be "independently useful" \cite{hinton2012improving}.


\subsection{Model Compression and Training Set Augmentation}
\label{sec:model_compression_and_training_set_augmentation}

As mentioned above in Section~\ref{sec:dropout}, ensembles of networks often achieve higher performance than any constituent network, but are computationally infeasible in most real-world applications. This is especially true for problems with small labeled datasets, where single networks are likely to overfit.

Bucilua et al. proposed that for semi-supervised problems, where a large dataset is available but only a small subset is labeled, it may be beneficial to train an ensemble (or another slow and accurate model) on the labeled subset, and use it to label all data, before finally training a single network (or another fast model) on the entire dataset, to reproduce both the original labels and the labels added by the ensemble \cite{bucilua2006model}.

In applications where there is a shortage of both labeled and unlabeled data, the training set can be artificially augmented. Training set augmentation is application-specific. For example, in computer vision applications where the network should be translationally and rotationally invariant, additional training entries can be formed by translating or rotating images from the original training set \cite{bengio2013representation}.


\subsection{Making Deep Nets Shallow}
\label{sec:making_deep_nets_shallow}

Continuing on the theme of model compression, in 2014, Ba and Caruana showed that with the help of a high performance deep network, a shallow network can be trained to perform much better than a similar shallow network that is trained directly on the training set \cite{ba2014deep}.

Their algorithm first trains a deep net (or an ensemble of deep nets) on the original training set, and use it to provide "extended labels" for all entries in the training set \cite{ba2014deep}. In case of classification problems where the output layer is often softmax, the "extended labels" are inputs to the softmax layer \cite{ba2014deep}. The inputs to the softmax layer are log probabilities of each class.

Finally, a shallow network can be trained to predict the log probabilities, instead of the original single-class label. This makes training much easier for the shallow network, because multi-class log probabilities labels provide much more information than a single-class label. By modelling the log probabilities, the shallow network is also mimic-ing how the deep net (or ensemble) will generalize to unseen data, which mostly depend on the relative values of log probabilities for classes that are not the highest \cite{ba2014deep}. The performances of these shallow networks are much higher than networks trained on the single-class labels.

This result is significant because it proves that the reason why shallow networks perform worse than deep networks is not entirely due to the increase in representative power and flexibility of deep networks. It is also due to our current training algorithms being sub-optimal for shallow networks, and if we can develop better training algorithms, we can potentially significantly improve the performance of shallow networks.

However, the performance of these mimic-ing shallow networks are still not quite as good as the deep networks or ensembles they are mimic-ing \cite{ba2014deep}. Therefore, the option of creating a mimic-ing shallow network allows a tradeoff to be made between accuracy and speed.


\subsection{Convolutional Neural Networks}
\label{sec:cnn}

Convolutional neural networks are a neural network architecture that uses extensive weight-sharing to reduce the degrees of freedom of models that operate on features that are spatially-correlated \cite{fukushima1980neocognitron}. This includes 2D and 3D images (and 2D videos, which can be seen as 3D images), but it has also very recently been successfully applied to natural language processing \cite{meng2015encoding}.

Convolutional neural networks are inspired by the observation that for inputs like images (with each pixel being an input dimension), many low level operations are local, and they are not position-dependent \cite{fukushima1980neocognitron}. For example, one operation that is useful in many computer vision applications is edge detection. In a fully-connected deep network, the edge detector would have to be separately trained for each part of the image, even though they would most likely all arrive at similar results. It would be better if a kernel can be trained to do edge detection for the entire image at the same time.

In its current iteration, convolutional neural networks are composed of 3 different types of layers - convolutional, max-pooling, and fully-connected \cite{krizhevsky2012imagenet}. One typical arrangement is alternating between convolutional and maxpooling layers, before finishing off with 2 fully-connected hidden layers.

Each convolutional layer has the same dimensions as the input, but each pixel is only activated by a region of pixels centered around the pixels at the same location in the input images. The weights are also shared for each output pixel. In effect, each map in a convolutional layer performs a convolution of the input images, with a learned kernel.

Max-pooling layers perform downsampling on the images. One typical downsample factor is 2x2 (dividing both width and height by 2). While averaging can also be used, empirical results suggest that downsampling by taking the maximum in each sub-region gives the best performance in most cases \cite{boureau2010theoretical}. Max-pooling is responsible for summarizing each sub-region, and it gives the network some translational and rotational invariance.

Fully-connected layers are often used as final layers to encode position-dependent information and more global patterns.

Most existing applications of convolutional neural networks are on 2D images, but the idea can also be extended to 3D, with 3D images and 3D kernels. It can be used to process actual 3D images (eg. MR Images), or videos, using time as the third dimension.



\section{Hippocampus Segmentation}
\label{sec:hippo}

\begin{figure}[ht!]
    \centering
    \includegraphics[scale=0.5]{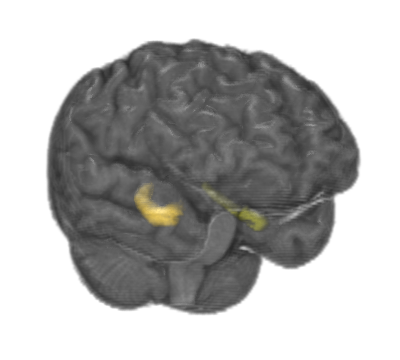}
    \caption{Hippocampus}
    \label{fig:hippo}
\end{figure}

The hippocampus is a component of human brains responsible for committing short-term episodic and declarative memory into long-term memory, as well as navigation \cite{tulving1998episodic}. Hippocampus segmentation is important in the diagnosis of Alzheimer's disease (AD), as it is one of the components first affected by the disease. A reduction in hippocampal volume can be used as a marker for AD diagnosis \cite{carmichael2005atlas}.

Humans have 2 hippocampi, shaped like seahorses, as shown in Figure~\ref{fig:hippo}. Our goal is to classify each voxel in an MR Image as non-hippocampus, left hippocampus, or right hippocampus. We are using this problem to evaluate different deep learning techniques for patch-based segmentation. All images are labeled by one human expert. Unfortunately, none of the images have been labeled by more than 1 human expert to determine variances in human labeling.

\subsection{Methodology}
\label{sec:methodology}

We explore 3 convolutional neural network architectures for patch-based segmentation on the ADNI Alzheimer's MRI dataset \cite{wyman2013standardization}.

For all 3 cases, the pre-processing and post-processing done are identical.

For all 3 cases, 60\% (120) of the images are used as the training set, 20\% (40) as the validation set, and 20\% (40) as the testing set. Patches from the same image are always only used in one of the sets.

\subsubsection{Pre-Processing}
\label{sec:preproc}

Before we begin labeling an image, we first crop it down to a rectangular bounding box, so we can perform masking in normalized coordinates. In case of the ADNI dataset, all images are already in the same orientation, so no rotation is needed.

From going through all images in the training set, we determined that the hippocampi are always in the region $(0.42 < x < 0.81, 0.30 < y < 0.67, 0.22 < z < 0.80)$, relative to each dimension of the bounding box of their respective brains. We enlarged the region by 0.03 on each side, and use $(0.39 < x < 0.84, 0.27 < y < 0.70, 0.19 < z < 0.83)$ as the mask. All voxels outside of the mask are automatically classified as non-hippocampus. All training patches are drawn from within the mask.


\subsubsection{Sampling}
\label{sec:sampling}

It would be dangerous to draw training voxels uniformly randomly from within the mask, because even within the mask, the vast majority of voxels are non-hippocampus, and hence there would be very few positive samples. Another problem is that edge voxels (voxels at the edges between positive and negative voxels) would be severely under-represented, even though they will most likely be the most difficult voxels to classify.

Therefore, we draw samples as follows -
\begin{itemize}
\item For 50\% of the samples, we keep drawing randomly until we get a voxel where the 5x5x5 bounding box around the voxel contains more than 1 class
\item For 25\% of the samples, we keep drawing randomly until we get a positive voxel
\item For the remaining 25\% of the samples, we keep drawing randomly until we get a negative voxel
\end{itemize}

This drawing scheme ensures that none of the important types of voxels are under-represented. The biggest downside of this scheme is that it distorts the prior probabilities of each class, possible solutions to which are discussed later in the report.


\subsubsection{Convolutional Method 1: Stacked 2D Patches}
\label{sec:2d_patches}

The first method we tried is to use a stack of 2D patches around each voxel we want to sample. For example, for a patch size of 24 and a layer count of 3, we would extract three 24x24 patches - one around the voxel in question, one in parallel and above, and one in parallel and below.

Each of the layers are given to a 2D convolutional neural network as different channels.

This method gives the network some 3D context around the voxel (in case of stack sizes greater than 1), at a relatively low space overhead. However, the network is not convolutional in the third dimension.

Network architecture is 20 5x5 kernels in first convolutional layer, 50 5x5 kernels in the second convolutional layer, a 1000 nodes fully-connected layer, then finally a softmax layer for exponential normalization. No max-pooling is used, since network performs slightly worse with any max-pooling.


\subsubsection{Convolutional Method 2: Tri-planar Patches}
\label{sec:trip_patches}

The second method we tried is the tri-planar method used by Prasoon et al. and Roth et al. in other medical imaging applications \cite{prasoon2013deep,roth2014new}.

For each voxel, we extract 3 square patches around the voxel, perpendicular to each axis. For example, if we want a patch size of 24, we would extract a 24x24 patch on the x-y plane centered around the voxel in question, and a 24x24 patch on the x-z plane, and another 24x24 patch on the y-z plane.

Since the corresponding pixels from the 3 patches are not spatially correlated in this case, we use a network architecture that consists of 2 convolutional layers (20 5x5 kernels and 50 5x5 kernels) for each of the 3 patches with no connections between them until the very end, where we feed all their outputs into a 1000 nodes fully-connected layer for final classification. No max-pooling is used.


\subsubsection{Convolutional Method 3: 3D Patches}
\label{sec:3d_patches}

This approach is an intuitive extension of the 2D approach into 3D. For each voxel we want to sample, we take a 3D patch with equal length on each side, around the voxel. For a patch size of 24, we would extract 24x24x24 patches.

Network architecture is 20 5x5x5 kernels in the first convolutional layer, 50 5x5x5 kernels in the second convolutional layer, then a 1000 nodes fully-connected layer as before. No max-pooling is used.


\subsubsection{Image Labeling}
\label{sec:labeling_images}

After the network is trained, to label an image, patches are extracted for every voxel in the mask region (in the correct format for the network architecture in use), and the result is used to label the voxel. Any voxel outside of the mask region is automatically classified as negative.


\subsubsection{Training}
\label{sec:training}

All network training are done with standard stochastic gradient descent with a batch size of 50 and a fixed learning rate of 0.01. 

At the beginning of training, termination iteration is set to 1 validation period. Validation is done after every pass through the training set (24,000 patches). Every time a validation score improves the current best validation score by more than 1\% (in error, not classification rate), the terminating iteration is set to twice the current iteration count. This means training will only be terminated if there is no significant improvement for at least the second half of the elapsed time.


\subsubsection{Post-Processing}
\label{sec:postproc}

\FloatBarrier
\begin{figure}[h]
\hfill
\subfigure[Before post-processing]{\includegraphics[scale=0.5]{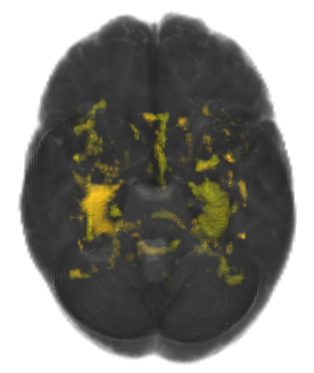}}
\hfill
\subfigure[After post-processing]{\includegraphics[scale=0.5]{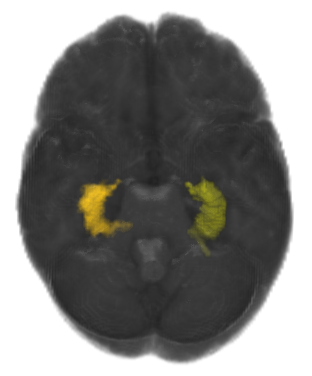}}
\hfill
\caption{Before and after post-processing}
\label{fig:before_after_postproc}
\end{figure}

Besides comparing raw output after labeling by convolutional neural networks, we also want to see what kind of performance can we get after some simple post-processing to clean up the results. The post-processing applied is the same for all 3 convolutional architectures.

For each labeled image, we first calculate the centroid of all voxels labeled left-hippocampus, and the centroid of all voxels labeled right-hippocampus.

Then, we divide up the image into blobs (connected voxels with the same classification), and for each blob, we check their size.

If a blob is smaller than a certain threshold (500 voxels in our case), and the labeling is negative (non-hippocampus), it is re-labeled to be the nearest hippocampus (based on centroid).

If a blob is smaller than the threshold, and the labeling is positive (hippocampus), it is re-labeled as negative (non-hippocampus).

We find that these simple post-processing steps clean up most of the obviously mis-classified voxels, as shown in Figure~\ref{fig:before_after_postproc}.



\FloatBarrier

\subsection{Results}
\label{sec:results}

All timing results in this section are obtained on a single NVIDIA GeForce GTX Titan Black GPU, using Theano's CUDA implementation of convolutional neural networks \cite{bergstra2010theano}.

Our first experiment is to determine the effect of number of layers on the performance of the 2D convolutional architecture. As shown in Table~\ref{tab:2d_num_layers}, there are very clear improvements going from 1 to 3 layers, but there are no clear improvements beyond 3 layers. However, we also note that number of layers has little impact on speed in terms of time per iteration.

All labeling results are on one of the images in the testing set, with 1755 positive voxels, and 1393947 negative voxels. All false positive and false negative values are after post-processing, and since the total number of positive and negative voxels are the same, the false positive and false negative values are directly comparable between runs. They are proportional to $(100\%-recall)$ and $(100\%-precision)$ respectively.

\FloatBarrier
\begin{table}[ht!]
\centering
\begin{tabular}{|l|l|l|l|l|l|l|}
\hline
\# Layers & Best Val Perf & Test Perf & False Pos (vxl) & False Neg (vxl) & Iter & T (mins) \\ \hline
1                & 21.55\%                     & 21.38\%             & 666                   & 426                   & 18240           & 22.15          \\ \hline
2                & 13.29\%                     & 14.56\%             & 1173                  & 413                   & 15360           & 27.67          \\ \hline
3                & 8.76\%                      & 9.79\%              & 513                   & 283                   & 25920           & 30.39          \\ \hline
4                & 11.69\%                     & 11.51\%             & 706                   & 446                   & 19680           & 25.33          \\ \hline
5                & 8.95\%                      & 9.64\%              & 490                   & 306                   & 18720           & 21.99          \\ \hline
7                & 9.34\%                      & 9.65\%              & 479                   & 312                   & 23040           & 27.65          \\ \hline
9                & 9.25\%                      & 10.03\%             & 577                   & 292                   & 22080           & 27.37          \\ \hline
11               & 10.04\%                     & 11.23\%             & 583                   & 331                   & 8640            & 20.37          \\ \hline
13               & 9.23\%                      & 10.01\%             & 761                   & 283                   & 28320           & 34.52          \\ \hline
\end{tabular}
\caption{Performance using different numbers of layers}
\label{tab:2d_num_layers}
\end{table}

The next experiment is to test different patch sizes, also with the 2D architecture. As shown in Table~\ref{tab:2d_patch_size}, there are little benefits in going beyond 24x24. However, in this case, training becomes much slower for larger patch sizes (in time per iteration). Therefore, the optimal patch size seems to be 24x24.

\FloatBarrier
\begin{table}[ht!]
\centering
\begin{tabular}{|l|l|l|l|l|l|l|}
\hline
Patch Size & Best Val Perf & Test Perf & False Pos (vxl) & False Neg (vxl) & Iter & T (mins) \\ \hline
12               & 11.33\%                     & 12.64\%             & 2341                  & 275                   & 72960           & 10.93          \\ \hline
16               & 10.20\%                     & 10.64\%             & 674                   & 234                   & 47520           & 12.41          \\ \hline
24               & 8.85\%                      & 10.00\%             & 778                   & 278                   & 24480           & 11.72          \\ \hline
32               & 8.98\%                      & 9.86\%              & 647                   & 283                   & 19680           & 13.72          \\ \hline
48               & 8.96\%                      & 9.80\%              & 547                   & 257                   & 22560           & 27.96          \\ \hline
\end{tabular}
\caption{Performance using different 2D patch sizes}
\label{tab:2d_patch_size}
\end{table}

Next, we experiment with patch size for the tri-planar architecture, and see similar results, with the optimal patch size being 24x24, as shown in Table~\ref{tab:trip_patch_size}. There are also very significant reductions in training speed as patch size is increased.

\FloatBarrier
\begin{table}[ht!]
\centering
\begin{tabular}{|l|l|l|l|l|l|l|}
\hline
Patch Size & Best Val Perf & Test Perf & False Pos (vxl) & False Neg (vxl) & Iter & T (mins) \\ \hline
12               & 28.74\%                     & 31.69\%             & 32446                 & 506                   & 6240            & 9.10           \\ \hline
16               & 8.74\%                      & 10.31\%             & 663                   & 251                   & 33120           & 96.84          \\ \hline
24               & 7.56\%                      & 8.29\%              & 775                   & 95                    & 29760           & 95.72          \\ \hline
32               & 7.23\%                      & 7.99\%              & 838                   & 118                   & 13920           & 95.58          \\ \hline
48               & 7.45\%                      & 8.45\%              & 626                   & 224                   & 2640            & 243.58         \\ \hline
\end{tabular}
\caption{Performance using different tri-planar patch sizes}
\label{tab:trip_patch_size}
\end{table}

Finally, we look at patch sizes for the 3D architecture. Unfortunately, we are constrained by available GPU memory in this case, and can only use up to 20x20x20 patches. We find that unlike the previous two architectures, the 3D architecture performs well even at a very small patch size of 12x12x12, and it is not clear whether it actually benefits from having larger patches, as shown in Table~\ref{tab:3d_patch_size}.

\FloatBarrier
\begin{table}[ht!]
\centering
\begin{tabular}{|l|l|l|l|l|l|l|}
\hline
Patch Size & Best Val Perf & Test Perf & False Pos (vxl) & False Neg (vxl) & Iter & T (mins) \\ \hline
12               & 7.54\%                      & 8.66\%              & 1690                  & 117                   & 12960           & 47.71          \\ \hline
16               & 6.73\%                      & 8.05\%              & 1799                  & 108                   & 7680            & 66.53          \\ \hline
20               & 7.06\%                      & 7.50\%              & 729                   & 127                   & 13920           & 194.08         \\ \hline
\end{tabular}
\caption{Performance using different 3D patch sizes}
\label{tab:3d_patch_size}
\end{table}

From the above experiments, we selected 3 configurations to investigate further -
\begin{itemize}
\item 2D 24x24, 7 layers
\item Tri-planar 24x24
\item 3D 20x20x20
\end{itemize}

We run each configuration five times to see how consistent they are. The networks are initialized with different random seeds each time, but trained on the same training sets. Results are presented in Table~\ref{tab:best_run}.

\FloatBarrier
\begin{table}[ht!]
\centering
\begin{tabular}{|l|l|l|l|l|l|l|}
\hline
Type & Best Val Perf & Test Perf & False Pos (vxl) & False Neg (vxl) & Iter & T (mins) \\ \hline
2D   & 8.18\%                      & 9.11\%              & 819                   & 178                   & 31680           & 42.06          \\ \hline
2D   & 8.80\%                      & 8.84\%              & 811                   & 235                   & 42240           & 51.56          \\ \hline
2D   & 8.76\%                      & 9.41\%              & 831                   & 207                   & 37920           & 49.36          \\ \hline
2D   & 8.03\%                      & 9.10\%              & 832                   & 172                   & 40800           & 50.21          \\ \hline
2D   & 8.61\%                      & 9.54\%              & 804                   & 221                   & 39840           & 44.64          \\ \hline
TriP & 7.59\%                      & 8.64\%              & 767                   & 120                   & 17280           & 78.00          \\ \hline
TriP & 7.64\%                      & 8.96\%              & 751                   & 107                   & 36480           & 141.55         \\ \hline
TriP & 7.35\%                      & 8.59\%              & 798                   & 106                   & 18720           & 78.14          \\ \hline
TriP & 7.39\%                      & 8.71\%              & 692                   & 127                   & 17280           & 100.38         \\ \hline
TriP & 7.33\%                      & 8.46\%              & 822                   & 102                   & 14880           & 80.51          \\ \hline
3D   & 6.95\%                      & 7.28\%              & 806                   & 113                   & 15360           & 161.67         \\ \hline
3D   & 6.69\%                      & 7.34\%              & 724                   & 136                   & 14880           & 187.57         \\ \hline
3D   & 7.46\%                      & 7.71\%              & 743                   & 127                   & 6720            & 81.50          \\ \hline
3D   & 6.58\%                      & 7.54\%              & 737                   & 130                   & 31680           & 301.54         \\ \hline
3D   & 7.14\%                      & 7.21\%              & 714                   & 155                   & 17760           & 173.73         \\ \hline
\end{tabular}
\caption{Multiple runs of best configurations with random initialization (2D 24x24 7 layers, tri-planar 24x24, 3D 20x20x20)}
\label{tab:best_run}
\end{table}

While the results on the validation and testing sets are reasonably consistent between each run of the same configuration, the actual labeling performance on an entire image is much less consistent. This can be due to the fact that the class distribution in the patches used for the training/validation/testing sets are not the same as the distribution in an actual image. The discrepancy in image labeling performance could be because some models do better at classifying some classes than others, and those classes are more common in an actual image.

From the perspective of speed-accuracy tradeoffs, the 2D architecture is clearly the least accurate, but also the fastest to train at approximately 754 iterations per minute.

The tri-planar architecture has clearly better performance than 2D, and can be trained at approximately 221 iterations per minute.

The 3D architecture performs consistently better than the tri-planar architecture in classifying patches, but that advantage does not seem to translate well to image labeling performance, where it seems to perform slightly worse. It is also the slowest to train at approximately 95 iterations per minute, with the longest run taking more than 5 hours to train.

\FloatBarrier



\section{Conclusion and Future Work}
\label{sec:conclusion}

In this project we investigated the use of three different convolutional network architectures for patch-based segmentation of the hippocampi region in MRI images. We discovered that the popular tri-planar approach offers a good tradeoff between accuracy and training time. While the 3D approach performs marginally better at patch classification, it does not seem to perform as well at labeling an entire image. This is most likely due to the sampling method altering prior probabilities of the classes presented to the training algorithm, and if this problem is solved, the 3D approach should perform marginally better than the tri-planar approach in whole-image labeling as well, but with a much higher computational power requirement.

There are many possible avenues for future investigation. For example, there are learning rate scheduling algorithms available that may significantly shorten training time without affecting quality of results, such as Zeiler's famous ADADELTA algorithm \cite{zeiler2012adadelta}, which assigns an independent learning rate to each weight depending on how often they are activated, so that while some oft-used connections can have slower learning rates to achieve higher precision, rarely-used connections can still be trained at a higher learning rate to reduce bias.

It may also be beneficial to give the coordinate of each patch (within the mask) to the fully-connected layers of the neural networks, along with the prior probability of each class at that coordinate, determined by statistical analysis on the training set.

One possible extension to the tri-planar architecture is to include images at multiple scales for each plane, similarly to how it was applied to traffic sign recognition by Sermanet and LeCun \cite{sermanet2011traffic}. This would allow the networks to have more global context, and if the largest scales are big enough to include boundaries of the brain, it may compensate for the lack of positional information in the tri-planar architecture. This can either be an alternative to, or complement, the idea of giving statistical coordinate-based prior probabilities to the network.

\newpage 

\bibliographystyle{unsrtnat}
	\bibliography{references} 

\begin{thebibliography}{36}
\providecommand{\natexlab}[1]{#1}
\providecommand{\url}[1]{\texttt{#1}}
\expandafter\ifx\csname urlstyle\endcsname\relax
  \providecommand{\doi}[1]{doi: #1}\else
  \providecommand{\doi}{doi: \begingroup \urlstyle{rm}\Url}\fi

\bibitem[Arel et~al.(2010)Arel, Rose, and Karnowski]{arel2010deep}
Itamar Arel, Derek~C Rose, and Thomas~P Karnowski.
\newblock Deep machine learning-a new frontier in artificial intelligence
  research [research frontier].
\newblock \emph{Computational Intelligence Magazine, IEEE}, 5\penalty0
  (4):\penalty0 13--18, 2010.

\bibitem[Krizhevsky et~al.(2012)Krizhevsky, Sutskever, and
  Hinton]{krizhevsky2012imagenet}
Alex Krizhevsky, Ilya Sutskever, and Geoffrey~E Hinton.
\newblock Imagenet classification with deep convolutional neural networks.
\newblock In \emph{Advances in neural information processing systems}, pages
  1097--1105, 2012.

\bibitem[Collobert and Weston(2008)]{collobert2008unified}
Ronan Collobert and Jason Weston.
\newblock A unified architecture for natural language processing: Deep neural
  networks with multitask learning.
\newblock In \emph{Proceedings of the 25th international conference on Machine
  learning}, pages 160--167. ACM, 2008.

\bibitem[Lee et~al.(2009)Lee, Pham, Largman, and Ng]{lee2009unsupervised}
Honglak Lee, Peter Pham, Yan Largman, and Andrew~Y Ng.
\newblock Unsupervised feature learning for audio classification using
  convolutional deep belief networks.
\newblock In \emph{Advances in neural information processing systems}, pages
  1096--1104, 2009.

\bibitem[Williams and Zipser(1989)]{williams1989learning}
Ronald~J Williams and David Zipser.
\newblock A learning algorithm for continually running fully recurrent neural
  networks.
\newblock \emph{Neural computation}, 1\penalty0 (2):\penalty0 270--280, 1989.

\bibitem[Leung et~al.(2003)Leung, Lam, Ling, and Tam]{leung2003tuning}
Frank Hung-Fat Leung, Hak-Keung Lam, Sai-Ho Ling, and Peter Kwong-Shun Tam.
\newblock Tuning of the structure and parameters of a neural network using an
  improved genetic algorithm.
\newblock \emph{Neural Networks, IEEE Transactions on}, 14\penalty0
  (1):\penalty0 79--88, 2003.

\bibitem[Belew et~al.(1990)Belew, McInerney, and
  Schraudolph]{belew1990evolving}
Richard~K Belew, John McInerney, and Nicol~N Schraudolph.
\newblock Evolving networks: Using the genetic algorithm with connectionist
  learning.
\newblock In \emph{In}. Citeseer, 1990.

\bibitem[Schaffer et~al.(1992)Schaffer, Whitley, and
  Eshelman]{schaffer1992combinations}
J~David Schaffer, Darrell Whitley, and Larry~J Eshelman.
\newblock Combinations of genetic algorithms and neural networks: A survey of
  the state of the art.
\newblock In \emph{Combinations of Genetic Algorithms and Neural Networks,
  1992., COGANN-92. International Workshop on}, pages 1--37. IEEE, 1992.

\bibitem[Schetinin(2003)]{schetinin2003learning}
Vitaly Schetinin.
\newblock A learning algorithm for evolving cascade neural networks.
\newblock \emph{Neural Processing Letters}, 17\penalty0 (1):\penalty0 21--31,
  2003.

\bibitem[Hornik et~al.(1989)Hornik, Stinchcombe, and
  White]{hornik1989multilayer}
Kurt Hornik, Maxwell Stinchcombe, and Halbert White.
\newblock Multilayer feedforward networks are universal approximators.
\newblock \emph{Neural networks}, 2\penalty0 (5):\penalty0 359--366, 1989.

\bibitem[Leshno et~al.(1993)Leshno, Lin, Pinkus, and
  Schocken]{leshno1993multilayer}
Moshe Leshno, Vladimir~Ya Lin, Allan Pinkus, and Shimon Schocken.
\newblock Multilayer feedforward networks with a nonpolynomial activation
  function can approximate any function.
\newblock \emph{Neural networks}, 6\penalty0 (6):\penalty0 861--867, 1993.

\bibitem[Glorot et~al.(2011)Glorot, Bordes, and Bengio]{glorot2011deep}
Xavier Glorot, Antoine Bordes, and Yoshua Bengio.
\newblock Deep sparse rectifier networks.
\newblock In \emph{Proceedings of the 14th International Conference on
  Artificial Intelligence and Statistics. JMLR W\&CP Volume}, volume~15, pages
  315--323, 2011.

\bibitem[Yam and Chow(2000)]{yam2000weight}
Jim~YF Yam and Tommy~WS Chow.
\newblock A weight initialization method for improving training speed in
  feedforward neural network.
\newblock \emph{Neurocomputing}, 30\penalty0 (1):\penalty0 219--232, 2000.

\bibitem[Pearlmutter(1991)]{pearlmutter1991gradient}
Barak~A Pearlmutter.
\newblock Gradient descent: Second-order momentum and saturating error.
\newblock \emph{Advances in neural information processing systems}, pages
  887--894, 1991.

\bibitem[Riedmiller and Braun(1992)]{riedmiller1992rprop}
Martin Riedmiller and Heinrich Braun.
\newblock Rprop-a fast adaptive learning algorithm.
\newblock In \emph{Proc. of ISCIS VII), Universitat}. Citeseer, 1992.

\bibitem[Schwarz et~al.(1978)]{schwarz1978estimating}
Gideon Schwarz et~al.
\newblock Estimating the dimension of a model.
\newblock \emph{The annals of statistics}, 6\penalty0 (2):\penalty0 461--464,
  1978.

\bibitem[Fukushima(1980)]{fukushima1980neocognitron}
Kunihiko Fukushima.
\newblock Neocognitron: A self-organizing neural network model for a mechanism
  of pattern recognition unaffected by shift in position.
\newblock \emph{Biological cybernetics}, 36\penalty0 (4):\penalty0 193--202,
  1980.

\bibitem[Hochreiter(1991)]{hochreiter1991untersuchungen}
Sepp Hochreiter.
\newblock Untersuchungen zu dynamischen neuronalen netzen.
\newblock \emph{Master's thesis, Institut fur Informatik, Technische
  Universitat, Munchen}, 1991.

\bibitem[Bengio et~al.(1994)Bengio, Simard, and Frasconi]{bengio1994learning}
Yoshua Bengio, Patrice Simard, and Paolo Frasconi.
\newblock Learning long-term dependencies with gradient descent is difficult.
\newblock \emph{Neural Networks, IEEE Transactions on}, 5\penalty0
  (2):\penalty0 157--166, 1994.

\bibitem[Hinton(2007)]{hinton2007learning}
Geoffrey~E Hinton.
\newblock Learning multiple layers of representation.
\newblock \emph{Trends in cognitive sciences}, 11\penalty0 (10):\penalty0
  428--434, 2007.

\bibitem[Vincent et~al.(2008)Vincent, Larochelle, Bengio, and
  Manzagol]{vincent2008extracting}
Pascal Vincent, Hugo Larochelle, Yoshua Bengio, and Pierre-Antoine Manzagol.
\newblock Extracting and composing robust features with denoising autoencoders.
\newblock In \emph{Proceedings of the 25th international conference on Machine
  learning}, pages 1096--1103. ACM, 2008.

\bibitem[Hinton et~al.(2012)Hinton, Srivastava, Krizhevsky, Sutskever, and
  Salakhutdinov]{hinton2012improving}
Geoffrey~E Hinton, Nitish Srivastava, Alex Krizhevsky, Ilya Sutskever, and
  Ruslan~R Salakhutdinov.
\newblock Improving neural networks by preventing co-adaptation of feature
  detectors.
\newblock \emph{arXiv preprint arXiv:1207.0580}, 2012.

\bibitem[Zhou et~al.(2002)Zhou, Wu, and Tang]{zhou2002ensembling}
Zhi-Hua Zhou, Jianxin Wu, and Wei Tang.
\newblock Ensembling neural networks: many could be better than all.
\newblock \emph{Artificial intelligence}, 137\penalty0 (1):\penalty0 239--263,
  2002.

\bibitem[Bucilua et~al.(2006)Bucilua, Caruana, and
  Niculescu-Mizil]{bucilua2006model}
Cristian Bucilua, Rich Caruana, and Alexandru Niculescu-Mizil.
\newblock Model compression.
\newblock In \emph{Proceedings of the 12th ACM SIGKDD international conference
  on Knowledge discovery and data mining}, pages 535--541. ACM, 2006.

\bibitem[Bengio et~al.(2013)Bengio, Courville, and
  Vincent]{bengio2013representation}
Yoshua Bengio, Aaron Courville, and Pascal Vincent.
\newblock Representation learning: A review and new perspectives.
\newblock \emph{Pattern Analysis and Machine Intelligence, IEEE Transactions
  on}, 35\penalty0 (8):\penalty0 1798--1828, 2013.

\bibitem[Ba and Caruana(2014)]{ba2014deep}
Jimmy Ba and Rich Caruana.
\newblock Do deep nets really need to be deep?
\newblock In \emph{Advances in Neural Information Processing Systems}, pages
  2654--2662, 2014.

\bibitem[Meng et~al.(2015)Meng, Lu, Wang, Li, Jiang, and Liu]{meng2015encoding}
Fandong Meng, Zhengdong Lu, Mingxuan Wang, Hang Li, Wenbin Jiang, and Qun Liu.
\newblock Encoding source language with convolutional neural network for
  machine translation.
\newblock \emph{arXiv preprint arXiv:1503.01838}, 2015.

\bibitem[Boureau et~al.(2010)Boureau, Ponce, and LeCun]{boureau2010theoretical}
Y-Lan Boureau, Jean Ponce, and Yann LeCun.
\newblock A theoretical analysis of feature pooling in visual recognition.
\newblock In \emph{Proceedings of the 27th International Conference on Machine
  Learning (ICML-10)}, pages 111--118, 2010.

\bibitem[Tulving and Markowitsch(1998)]{tulving1998episodic}
Endel Tulving and Hans~J Markowitsch.
\newblock Episodic and declarative memory: role of the hippocampus.
\newblock \emph{Hippocampus}, 8\penalty0 (3), 1998.

\bibitem[Carmichael et~al.(2005)Carmichael, Aizenstein, Davis, Becker,
  Thompson, Meltzer, and Liu]{carmichael2005atlas}
Owen~T Carmichael, Howard~A Aizenstein, Simon~W Davis, James~T Becker, Paul~M
  Thompson, Carolyn~Cidis Meltzer, and Yanxi Liu.
\newblock Atlas-based hippocampus segmentation in alzheimer's disease and mild
  cognitive impairment.
\newblock \emph{Neuroimage}, 27\penalty0 (4):\penalty0 979--990, 2005.

\bibitem[Wyman et~al.(2013)Wyman, Harvey, Crawford, Bernstein, Carmichael,
  Cole, Crane, DeCarli, Fox, Gunter, et~al.]{wyman2013standardization}
Bradley~T Wyman, Danielle~J Harvey, Karen Crawford, Matt~A Bernstein, Owen
  Carmichael, Patricia~E Cole, Paul~K Crane, Charles DeCarli, Nick~C Fox,
  Jeffrey~L Gunter, et~al.
\newblock Standardization of analysis sets for reporting results from adni mri
  data.
\newblock \emph{Alzheimer's \& Dementia}, 9\penalty0 (3):\penalty0 332--337,
  2013.

\bibitem[Prasoon et~al.(2013)Prasoon, Petersen, Igel, Lauze, Dam, and
  Nielsen]{prasoon2013deep}
Adhish Prasoon, Kersten Petersen, Christian Igel, Fran{\c{c}}ois Lauze, Erik
  Dam, and Mads Nielsen.
\newblock Deep feature learning for knee cartilage segmentation using a
  triplanar convolutional neural network.
\newblock In \emph{Medical Image Computing and Computer-Assisted
  Intervention--MICCAI 2013}, pages 246--253. Springer, 2013.

\bibitem[Roth et~al.(2014)Roth, Lu, Seff, Cherry, Hoffman, Wang, Liu, Turkbey,
  and Summers]{roth2014new}
Holger~R Roth, Le~Lu, Ari Seff, Kevin~M Cherry, Joanne Hoffman, Shijun Wang,
  Jiamin Liu, Evrim Turkbey, and Ronald~M Summers.
\newblock A new 2.5 d representation for lymph node detection using random sets
  of deep convolutional neural network observations.
\newblock In \emph{Medical Image Computing and Computer-Assisted
  Intervention--MICCAI 2014}, pages 520--527. Springer, 2014.

\bibitem[Bergstra et~al.(2010)Bergstra, Breuleux, Bastien, Lamblin, Pascanu,
  Desjardins, Turian, Warde-Farley, and Bengio]{bergstra2010theano}
James Bergstra, Olivier Breuleux, Fr{\'e}d{\'e}ric Bastien, Pascal Lamblin,
  Razvan Pascanu, Guillaume Desjardins, Joseph Turian, David Warde-Farley, and
  Yoshua Bengio.
\newblock Theano: a cpu and gpu math expression compiler.
\newblock In \emph{Proceedings of the Python for scientific computing
  conference (SciPy)}, volume~4, page~3. Austin, TX, 2010.

\bibitem[Zeiler(2012)]{zeiler2012adadelta}
Matthew~D Zeiler.
\newblock Adadelta: an adaptive learning rate method.
\newblock \emph{arXiv preprint arXiv:1212.5701}, 2012.

\bibitem[Sermanet and LeCun(2011)]{sermanet2011traffic}
Pierre Sermanet and Yann LeCun.
\newblock Traffic sign recognition with multi-scale convolutional networks.
\newblock In \emph{Neural Networks (IJCNN), The 2011 International Joint
  Conference on}, pages 2809--2813. IEEE, 2011.

\end{thebibliography}

\end{document}